# Automatic extraction of paraphrastic phrases from medium size corpora

**Thierry Poibeau**

Laboratoire d'Informatique de Paris-Nord – CNRS UMR 7030

Av. J.B. Clément – F-93430 Villetaneuse

thierry.poibeau@lipn.univ-paris13.fr

## Abstract

This paper presents a versatile system intended to acquire paraphrastic phrases from a representative corpus. In order to decrease the time spent on the elaboration of resources for NLP system (for example Information Extraction, IE hereafter), we suggest to use a knowledge acquisition module that helps extracting new information despite linguistic variation (textual entailment). This knowledge is automatically derived from the text collection, in interaction with a large semantic network.

## 1 Introduction

Recent researches in NLP have promoted a now widely-accepted shallow-based analysis framework that has proven to be efficient for a number of tasks, including information extraction and question answering. However, this approach often leads to over-simplified solutions to complex problems. For example, the bag-of-words approach fails in examples such as: *Lee Harvey Oswald, the gunman who assassinated President John F. Kennedy, was later shot and killed by Jack Ruby* (example taken from Lin and Katz, 2003). In this case, it is essential to keep track of the argument structure of the verb, to be able to infer that it is *Jack Ruby* and not *John Kennedy* who is the murderer of *Lee Harvey Oswald*. A wrong result would be obtained considering too shallow analysis techniques or heuristics, based for example of the proximity between two person names in the sentence.

Several studies have recently proposed some approaches based on the redundancy of the web to acquire extraction patterns and semantic structures. However, these methods cannot be applied to medium size corpora. Moreover, existing structured knowledge contained in dictionaries, thesauri or semantic networks can boost the learning process by providing clear intuition over text units.

In this paper, we propose a knowledge rich approach to paraphrase acquisition. We will firstly describe some related work for the acquisition of knowledge, especially paraphrases, from texts. We then describe how semantic similarity between words can be inferred from large semantic networks. We present an acquisition process, in which the semantic network is projected on the corpus to derive extraction patterns. This mechanism can be seen as a dynamic lexical tuning of information contained in the semantic network in order to generate paraphrases of an original pattern. In the last section, we propose an evaluation and some perspectives.

## 2 Related work

This section presents some related works for the acquisition of extraction patterns and paraphrases from texts.

### 2.1 IE and resource acquisition

IE is known to have established a now widely accepted linguistic architecture based on cascading automata and domain-specific knowledge (Appelt *et al*, 1993). However, several studies have outlined the problem of the definition of the resources. For example, E. Riloff (1995) says that about 1500 hours are necessary to define the resources for a text classification system on terrorism[1]. Most of these resources are variants of extraction patterns, which have to be manually established.

---

[1] We estimate that the development of resources for IE is at least as long as for text classification.

To address this problem of portability, a recent research effort focused on using machine learning throughout the IE process (Muslea, 1999). A first trend was to directly apply machine learning methods to replace IE components. For example, statistical methods have been successfully applied to the named-entity task. Among others, (Bikel *et a.*, 1997) learns names by using a variant of hidden Markov models.

## 2.2 Extraction pattern learning

Another research area trying to avoid the time-consuming task of elaborating IE resources is concerned with the generalization of extraction patterns from examples. (Muslea, 1999) gives an extensive description of the different approaches of that problem. Autoslog (Riloff, 1993) was one of the very first systems using a simple form of learning to build a dictionary of extraction patterns. Ciravegna (2001) demonstrates the interest of independent acquisition of left and right boundaries of extraction patterns during the learning phase. In general, the left part of a pattern is easier to acquire than the right part and some heuristics can be applied to infer the right boundary from the left one. The same method can be applied for argument acquisition: each argument can be acquired independently from the others since the argument structure of a predicate in context is rarely complete.

Collins and Singer (1999) demonstrate how two classifiers operating on disjoint features sets recognize named entities with very little supervision. The method is interesting in that the analyst only needs to provide some seed examples to the system in order to learn relevant information. However, these classifiers must be made interactive in order not to diverge from the expected result, since each error is transmitted and amplified by subsequent processing stages. Contrary to this approach, partially reproduced by Duclaye *et al.* (2003) for paraphrase learning, we prefer a slightly supervised method with clear interaction steps with the analyst during the acquisition process, to ensure the solution is converging.

## 3 Overview of the approach

Argument structure acquisition is a complex task since the argument structure is rarely complete. To overcome this problem, we propose an acquisition process in which all the arguments are acquired separately.

Figure 1 presents an outline of the overall paraphrase acquisition strategy. The process is made of automatic steps and manual validation stages. The process is weakly supervised since the analyst only has to provide one example to the system. However, we observed that the quality of the acquisition process highly depends from this seed example, so that several experiments has to be done for the acquisition of an argument structure, in order to be sure to obtain an accurate coverage of a domain.

From the seed pattern, a set of paraphrases is automatically acquired, using similarity measures between words and a shallow syntactic analysis of the found patterns, in order to ensure they describe a predicative sequence. All these stages are described below, after the description of similarity measures allowing to calculate the semantic proximity between words.

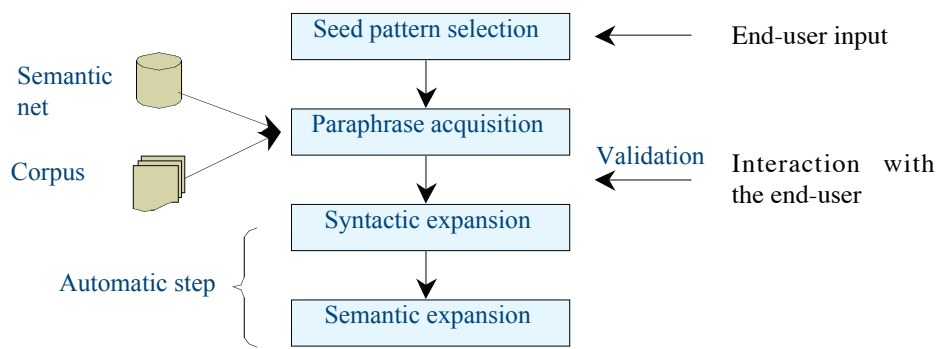

**Figure 1:** Outline of the acquisition process

## 4 Similarity measures

Several studies have recently proposed measures to calculate the semantic proximity between words. Different measures have been proposed, which are not easy to evaluate (see (Lin and Pantel, 2002) for proposals). The methods proposed so far are automatic or manual and generally imply the evaluation of word clusters in different contexts (a word cluster is close to another one if the words it contains are interchangeable in some linguistic contexts).

Budanitsky and Hirst (2001) present the evaluation of 5 similarity measures based on the structure of Wordnet. All the algorithms they examine are based on the hypernym-hyponym relation which structures the classification of clusters inside Wordnet (the *synsets*). They sometimes obtain unclear conclusions about the reason of the performances of the different algorithms (for example, comparing Jiang and Conrath's measure (1997) with Lin's one (1998): "It remains unclear, however, just why it performed so much better than Lin's measure, which is but a different arithmetic combination of the same terms"). However, the authors emphases on the fact that the use of the sole hyponym relation is insufficient to capture the complexity of meaning: "Nonetheless, it remains a strong intuition that hyponymy is only one part of semantic relatedness; meronymy, such as *wheel–ca*r, is most definitely an indicator of semantic relatedness, and, *a fortior*i, semantic relatedness can arise from little more than common or stereotypical associations or statistical co-occurrence in real life (for example, *penguin–Antarctica; birthday–candle; sleep–pajama*s)".

In this paper, we propose to use the semantic distance described in (Dutoit *et al.*, 2002) which is based on a knowledge-rich semantic net encoding a large variety of semantic relationships between set of words, including meronymy and stereotypical associations.

The semantic distance between two words A and B is based on the notion of nearest common ancestors (NCA) between A and B. NCA is defined as the set of nodes that are daughters of c(A) ∩ c(B) and that are not ancestors in c(A) ∩ c(B). The *activation measure* d_ is equal to the mean of the weight of each NCA calculated from A and B :

$$d_{\curlywedge}(A, B) = \frac{1}{n}\sum_{i=1}^{n}(d(A, NCA_i) + d(B, NCA_i))$$

Please, refer to (Dutoit and Poibeau, 2002) for more details and examples. However, this measure is sensitive enough to give valuable results for a wide variety of applications, including text filtering and information extraction (Poibeau *et al.*, 2002).

## 5 The acquisition process

The process begins as the end-user provides a predicative linguistic structure to the system along with a representative corpus. The system tries to discover relevant parts of text in the corpus based on the presence of plain words closely related to the ones of the seed pattern. A syntactic analysis of the sentence is then done to verify that these plain words correspond to a paraphrastic structure. The method is close to the one of Morin and Jacquemin (1999), who first try to locate couples of relevant terms and then apply relevant patterns to analyse the nature of their relationship. However, Morin and Jacquemin only focus on term variations whereas we are interested in predicative structures, being either verbal or nominal. The syntactic variations we have to deal with are then different and, for a part, more complex than the ones examined by Morin and Jacquemin.

The detail algorithm is described below:

1. The head noun of the example pattern is compared with the head noun of the candidate pattern using the proximity measure from (Dutoit *et al.*, 2002). This result of the measure must be under a threshold fixed by the end-user.

2. The same condition must be filled by the "expansion" element (possessive phrase or verb complement in the candidate pattern).

3. The structure must be predicative (either a nominal or a verbal predicate, the algorithm does not make any difference at this level).

The following schema (Figure 2) resumes the acquisition process.

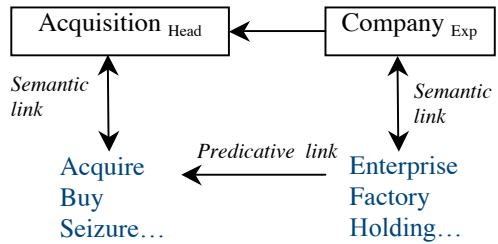

**Figure 2:** paraphrase acquisition

Finally, this process is formalized throughout the algorithm 1. Note that the predicative form is acquired together with its arguments, as in a co-training process.

```
P ← pattern to be found
S ← Sentence to analyze
C ← Phrases(S)
W ← Plain_words(S)
Result ← empty list
head ← Head word of the pattern P
exp ← Expansion word of the pattern P
Threshold ← threshold fixed by the analyst
For every word wᵢ from W do
 Prox₁ = d'⊥(head, wᵢ)
 If (Prox₁ <= Threshold) then
   wᵢ₊₁ ← Next element from W (if end of sentence then exit)
   Prox₂ = d'⊥(exp, wᵢ₊₁)
   If (Prox₂ <= Threshold) then
     If there is c ∈ C so that (wᵢ ∈ c) and (wᵢ₊₁ ∈ c) then
       Result ← Add (wᵢ, wᵢ₊₁)
     End_if
   End_if
 End_if
End_for
```

**Algorithm 1 :** Paraphrastic phrases acquisition

The result of this analysis is a table representing predicative structures, which are semantically equivalent to the initial example pattern. The process uses the corpus and the semantic net as two different complementary knowledge sources:
- The semantic net provides information about lexical semantics and relations between words
- The corpus attests possible expressions and filter irrelevant ones.

We performed an evaluation on different French corpora, given that the semantic net is especially rich for this language. We take the expression *cession de société* (*company transfer*) as an initial pattern. The system then discovered the following expressions, each of them being semantic paraphrases of the initial seed pattern:

```
reprise des activités
rachat d'activité
acquérir des magasins
racheter *c-company*
cession de *c-company*…
```

The result must be manually validated. Some structures are found even if they are irrelevant, due to the activation of irrelevant links. It is the case of the expression *renoncer à se porter acquéreur* (*to give up buying sthg*), which is not relevant. In this case, there was a spurious link between *to give up* and *company* in the semantic net.

### 5.1 Dealing with syntactic variations

The previous step extract semantically related predicative structures from a corpus. These structures are found in the corpus in various linguistic structures, but we want the system to be able to find this information even if it appears in other kind of linguistic sequences. That is the reason why we associate some meta-graphs with the linguistic structures, so that different transformations can be recognized. This strategy is based on Harris theory of sublanguages (1991). These transformations concern the syntactic level, either on the head (H) or on the expansion part (E) of the linguistic structure.

![Figure 3 spreadsheet screenshot]

**Figure 3:** the linguistic constraint table

The meta-graphs encode transformations concerning the following structures:

- Subject — verb,
- Verb — direct object,
- Verb — indirect object (especially when introduced by the French preposition *à* or *de*),
- Noun — possessive phrase.

These meta-graphs encode the major part of the linguistic structures we are concern with in the process of IE.

The graph on Figure 4 recognizes the following sequences (in brackets we underline the couple of words previously extracted from the corpus):

```
Reprise des activités charter… (H:
  reprise, E: activité)
Reprendre les activités charter…
  (H: reprendre, E: activité)
Reprise de l'ensemble des magasins
  suisse… (H: reprise, E: magasin)
Reprendre l'ensemble des magasins
  suisse… (H: reprendre, E: magasin)
Racheter les différentes activités…
  (H: racheter, E: activité)
Rachat des différentes activités…
  (H: rachat, E: activité)
```

This kind of graph is not easy to read. It includes at the same time some linguistic tags and some applicability constraints. For example, the first box contains a reference to the `@A` column in the table of identified structures. This column contains a set of binary constraints, expressed by some signs + or -. The sign + means that the identified pattern is of type verb-direct object: the graph can then be applied to deal with passive structures. In other words, the graph can only be applied in a sign + appears in the `@A` column of the constraints table. The constraints are removed from the instantiated graph. Even if the resulting graph is normally not visible (the compilation process directly produced a graph in a binary format), we give an image of a part of that graph on Figure 4.

This mechanism using constraint tables and meta-graph has been implemented in the finite-state toolbox INTEX (Silberztein, 1993). 26 meta-graphs have been defined modeling linguistic variation for the 4 predicative structures defined above. The phenomena mainly concern the insertion of modifiers (with the noun or the verb), verbal transformations (passive) and phrasal structures (relative clauses like *…Vivendi, qui a racheté Universal…Vivendi, that bought Universal*).

The compilation of the set of meta-graphs produces a graph made of 317 states and 526

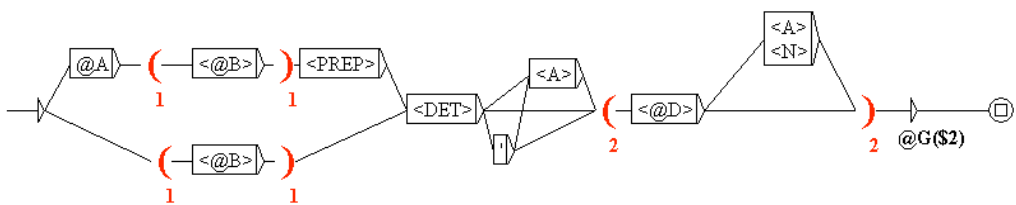

**Figure 4:** a syntactic meta-graph

relations. These graphs are relatively abstract but the end-user is not intended to directly manipulate them. They generate instantiated graphs, that is to say graphs in which the abstract variables have been replaced linguistic information as modeled in the constraint tables. This method associates a couple of elements with a set of transformation that covers more examples than the one of the training corpus. This generalization process is close to the one imagined by Morin and Jacquemin (1999) for terminology analysis but, as we already said, we cover sequences that are not only nominal ones.

## 6 Evaluation

The evaluation concerned the extraction of information from a French financial corpus, about companies buying other companies. The corpus is made of 300 texts (200 texts for the training corpus, 100 texts for the test corpus).

A system was first manually developed and evaluated. We then tried to perform the same task with automatically developed resources, so that a comparison is possible. The corpus is firstly normalized. For example, all the company names are replaced by a variable *c-company* thanks to the named entity recognizer. In the semantic network, *c-company* is introduced as a synonym of company, so that all the sequences with a proper name corresponding to a company could be extracted.

For the slot corresponding to the company that is being bought, 6 seed example patterns were given to semantic expansion module. This module acquired from the corpus 25 new validated patterns. Each example pattern generated 4.16 new patterns on average. For example, from the pattern rachat de *c-company* we obtain the following list:

```
reprise de *c-company*
achat de *c-company*
acquérir *c-company*
racheter *c-company*
cession de *c-company*
```

This set of paraphrastic patterns includes nominal phrases (reprise de *c-company*) and verbal phrases (racheter *c-company*). The acquisition process concerns at the same time, the head and the expansion. The simultaneous acquisition of different semantic classes can also be found in the co-training algorithm proposed for this kind of task by E. Riloff and R. Jones (Riloff et Jones, 1999).

The proposed patterns must be filtered and validated by the end-user. We estimate that generally 25% of the acquired pattern should be rejected. However, this validation process is very rapid: a few minutes only were necessary to check the 31 proposed patterns and retain 25 of them.

We then compared these results with the ones obtained with the manually elaborated system. The evaluation concerned the three slots that necessitate a syntactic and semantic analysis: the company that is buying another one (arg1) the company that is being bought (arg2), the company that sells (arg3). These slots imply nominal phrases, they can be complex and a functional analysis is most of the time necessary (is the nominal phrase the subject or the direct object of the sentence?). We thus chose to perform an operational evaluation: what is evaluated is the ability of a given phrase or pattern to fill a given slot (also called textual entailment by Dagan and Glickman [2004]). This kind of evaluation avoids, as far as possible, the bias of human judgment on possibly ambiguous expressions.

An overview of the results is given below (P refers to precision, R to recall, F to the harmonic mean between P and R):

|  | Arg 1 | Arg 2 | Arg 3 |
|---|---|---|---|
| **Human annotators** | P: 100 R: 90 | P: 100 R: 91.6 | P: 99 R: 92 |
|  | F: **94.7** | F: **95.6** | F: **94.2** |
| **Automatically acquired resources** | P: 79.6 R: 62.6 | P: 93.4 R: 73 | P: 88.4 R: 70 |
|  | F: **70** | F: **81.9** | F: **77** |

We observed that the system running with automatically defined resources is about 10% less efficient than the one with manually defined resources. The decrease of performance may vary in function of the slot (the decrease is less important for the arg2 than for arg1 or arg3). Two kind of errors are observed: Certain sequences are not found because a relation between words is missing in the semantic net. Some sequences are extracted by the semantic analysis but do not correspond to a transformation registered in the syntactic variation management module.

# 7 Conclusion

In this paper, we have shown an efficient algorithm to semi-automatically acquire paraphrastic phrases from a semantic net and a corpus. We have shown that this approach is highly relevant in the framework of IE systems. Even if the performance decrease when the resources are automatically defined, the gain in terms of development time is sufficiently significant to ensure the usability of the method.

# 8 References

Appelt D.E, Hobbs J., Bear J., Israel D., Kameyana M. and Tyson M. (1993) FASTUS: a finite-state processor for information extraction from real-world text. Proceedings of IJCAI'93, Chambéry, France, pp. 1172—1178.

Bikel D., Miller S., Schwartz R. and Weischedel R. (1997) Nymble: a high performance learning name-finder. Proceeding of the 5th ANLP Conference, Washington, USA.

Budanitsky A. and Hirst G. (2001) Semantic distance in WordNet: An experimental, application-oriented evaluation of five measures. Workshop on WordNet and Other Lexical Resources, in NAACL 2001, Pittsburgh.

Ciravegna F. (2001) Adaptive Information Extraction from Text by Rule Induction and Generalisation. Proceedings of the 17th International Joint Conference on Artificial Intelligence (IJCAI'2001), Seattle, pp. 1251–1256.

Collins M. and Singer Y. (1999) Unsupervised models for named entity classification. Proceedings of EMNLP-WVLC'99, College Park, pp. 100–110.

Dagan I. and Glickman O. (2004) Probabilistic Textual Entailment: Generic Applied Modeling of Language Variability. Workshop Learning Methods for Text Understanding and Mining. Grenoble, France.

Duclaye F., Yvon F. and Collin O. (2003) Learning paraphrases to improve a question answering system. Proceeding of the EACL Workshop "NLP for Question Answering", Budapest, Hungary.

Dutoit D. and Poibeau T. (2002) Deriving knowledge from a large semantic network, Proceedings of COLING'2002, Taipei, Taiwan, pp. 232—238.

Fellbaum C. (1998) WordNet : An Electronic Lexical Database, edited by Fellbaum, MIT press.

Grefenstette G. (1998) Evaluating the adequancy of a multilingual transfer dictionary for the Cross Language Information Retrieval, LREC 1998.

Harris Z. (1991) *A theory of language and information: a mathematical approach.* Oxford University Press. Oxford.

Jiang J. and Conrath D. (1997) Semantic similarity based on corpus statistics and lexical taxonomy. Proceedings of International Conference on Research in Computational Linguistics, Taiwan.

Jones R., McCallum A., Nigam K. and Riloff E. (1999) Bootstrapping for Text Learning Tasks. Proceedings of the IJCAI'99 Workshop on Text Mining: Foundations, Techniques and Applications, Stockholm, 1999, pp. 52—63.

Lin D. (1998) An information-theoretic definition of similarity. Proceedings of the 15th International Conference on Machine Learning, Madison, WI.

Lin D. and Pantel P. (2002) Concept Discovery from Text. Proceedings of COLING'2002, Taipei, Taiwan, pp. 577—583.

Lin J. and Katz B. (2003) Q/A techniques for WWW. Tutorial. 10th Meeting of the European Association for Computational Linguistics (EACL'03), Budapest, 2003.

Morin E. and Jacquemin C. (1999) Projecting corpus-based semantic links on a thesaurus. Proceedings of the 37th ACL, pp. 389–396.

Muslea I. (1999) Extraction patterns for Information Extraction tasks: a survey, AAAI'99 (available at the following URL: http://www.isi.edu/~muslea/ RISE/ML4IE/)

Pazienza M.T, ed. (1997) Information extraction. Springer Verlag (Lecture Notes in computer Science), Heidelberg, Germany.

Poibeau T., Dutoit D., Bizouard S. (2002) Evaluating resource acquisition tools for Information Extraction. Proceeding of the *International Language Resource and Evaluation Conference* (LREC 2002), Las Palmas.

Riloff E. (1993) Automatically constructing a dictionary for formation extraction tasks, AAAI'93, Stanford, USA, pp. 811—816.

Riloff E. (1995) Little Words Can Make a Big Difference for Text Classification, Proceedings of the SIGIR'95, Seattle, USA, pp. 130—136.

Riloff E. et Jones R. (1999) Learning Dictionaries for Information Extraction by Multi-Level Bootstrapping. Proceedings of the 16th National Conference on Artificial Intelligence (AAAI'99), Orlando, 1999, pp. 474—479.

Silberztein M. (1993) *Dictionnaires électroniques et analyse automatique des textes*, Masson, Paris, France.